%% file: 1408-arxiv.tex
\documentclass[runningheads]{llncs}
\usepackage{graphicx}
\usepackage{amsmath,amssymb}
\usepackage{color}
\usepackage{microtype}
\usepackage{epsfig}
\usepackage{caption}
\usepackage{subcaption}
\usepackage{fathy-defs}
\usepackage{microtype}
\usepackage{arydshln}
\usepackage{enumitem}
\setitemize[0]{leftmargin=10pt}
\usepackage{xspace}

\newenvironment{tight_itemize}{
\begin{itemize}[leftmargin=10pt]
  \setlength{\topsep}{0pt}
  \setlength{\itemsep}{2pt}
  \setlength{\parskip}{0pt}
  \setlength{\parsep}{0pt}
}{\end{itemize}}

\makeatletter
\DeclareRobustCommand\onedot{\futurelet\@let@token\@onedot}
\def\@onedot{\ifx\@let@token.\else.\null\fi\xspace}

\makeatletter
\def\blfootnote{\xdef\@thefnmark{}\@footnotetext}
\makeatother

\def\eg{\emph{e.g}\onedot} 
\def\ie{\emph{i.e}\onedot}

\makeatother

\begin{document}

\title{Hierarchical Metric Learning and Matching for 2D and 3D Geometric Correspondences$^{\star}$}

\titlerunning{Hierarchical Metric Learning and Matching for Geometric Correspondences}

\author{Mohammed E. Fathy$^1$ \and Quoc-Huy Tran$^2$ \and M. Zeeshan Zia$^3$ \and Paul Vernaza$^2$ \and Manmohan Chandraker$^{2,4}$}
\institute{$^1$Google Cloud AI \hfill $^3$Microsoft Hololens\\ $^2$NEC Laboratories America, Inc. \hfill $^4$University of California, San Diego}

\authorrunning{M. E. Fathy, Q.-H. Tran, M. Z. Zia, P. Vernaza, and M. Chandraker}

\maketitle

\input{abstract}

\input{introduction}

\input{related}

\input{method}

\input{experiment}

\input{conclusion}

\bibliographystyle{splncs04}
\bibliography{1408-arxiv}
\end{document}

%% file: abstract.tex
\begin{abstract}
Interest point descriptors have fueled progress on almost every problem in computer vision. Recent advances in deep neural networks
 have enabled task-specific learned descriptors that outperform hand-crafted descriptors on many problems. We demonstrate that commonly 
 used metric learning approaches do not optimally leverage the feature hierarchies learned in a Convolutional Neural Network (CNN), 
 especially when applied to the task of geometric feature matching. While a metric loss applied to the deepest layer of a CNN, is often 
 expected to yield ideal features irrespective of the task, in fact the growing receptive field as well as striding effects cause shallower
  features to be better at high precision matching tasks.  We leverage this insight together with explicit supervision at multiple 
  levels of the feature hierarchy for better regularization, to learn more effective descriptors in the context of geometric matching 
  tasks. Further, we propose to use activation maps at different layers of a CNN, as an effective and principled replacement for the 
  multi-resolution image pyramids often used for matching tasks. We propose concrete CNN architectures employing these ideas, and evaluate 
  them on multiple datasets for 2D and 3D geometric matching as well as optical flow, demonstrating state-of-the-art results and generalization across datasets.

\keywords{Hierarchical metric learning \and Hierarchical matching \and geometric correspondences \and dense correspondences}
\end{abstract}

%% file: introduction.tex
\section{Introduction}
{\blfootnote{{$^{\star}$Part of this work was done during M. E. Fathy's internship at NEC Labs America. Code and models will be made available at \textcolor{blue}{http://www.nec-labs.com/$\sim$mas/HiLM/}.}}}
The advent of repeatable high curvature point detectors~\cite{harris1988avc,lindeberg1998ijcv,lowe2004ijcv} heralded a revolution in computer vision that shifted the emphasis of the field from holistic object models and direct matching of image patches~\cite{zhang1995ai}, to highly discriminative hand-crafted descriptors. These descriptors made a mark on a wide array of problems in computer vision, with pipelines created to solve tasks such as optical flow~\cite{brox2011pami}, object detection~\cite{dalal2005cvpr}, 3D reconstruction~\cite{snavely2006tog} and action recognition~\cite{wang2011cvpr}.

\begin{figure}[!!t]
  \centering
  \begin{minipage}[c]{0.6\textwidth}
    \includegraphics[width=0.98\linewidth]{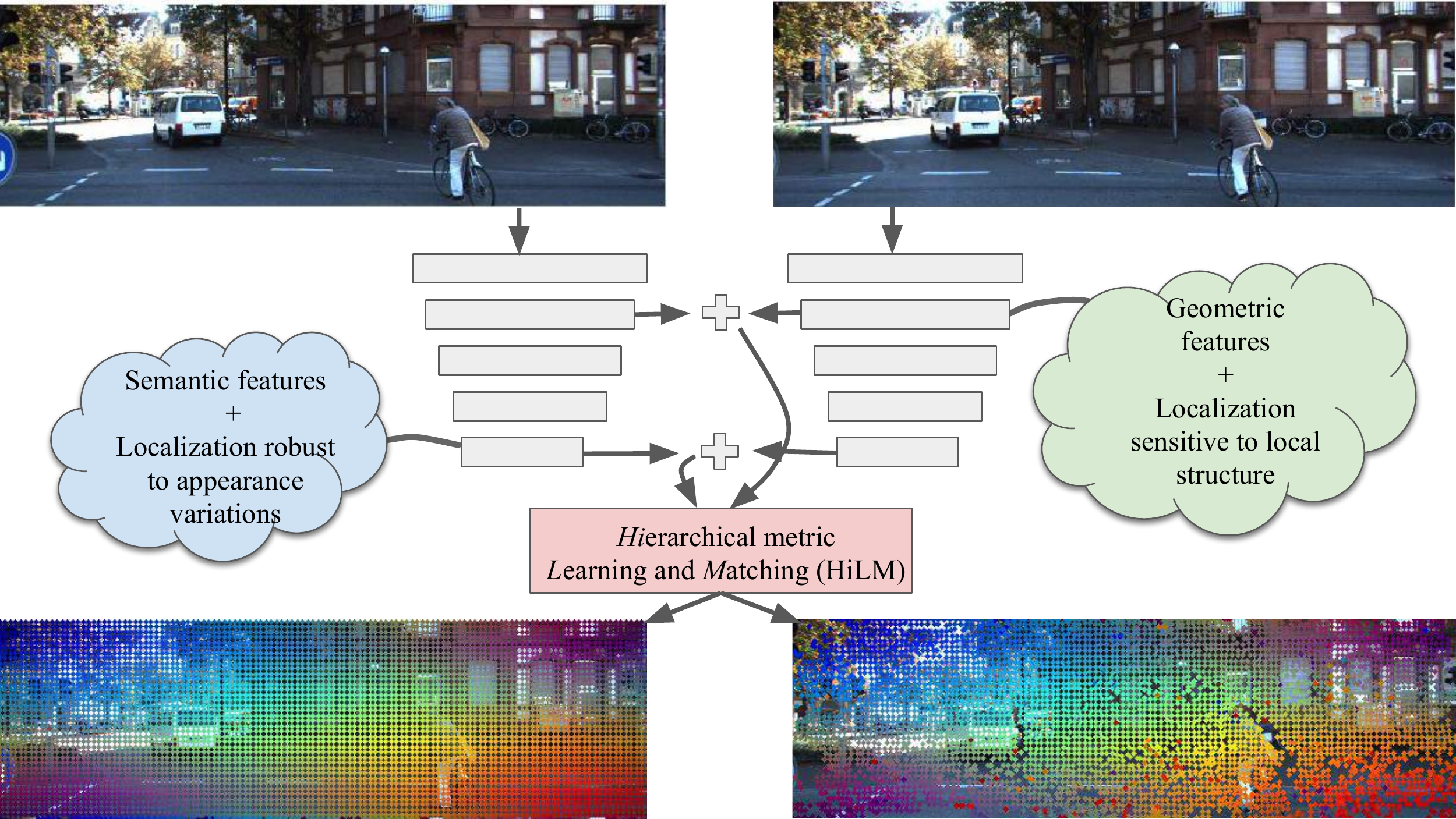}
  \end{minipage}\hfill
  \begin{minipage}[c]{0.4\textwidth}    
    \caption{\small
      Our hierarchical metric learning retains the best properties of various levels of abstraction in CNN feature representations. For geometric matching, we combine the robustness of deep layers that imbibe greater invariance, with the localization sensitivity of shallow layers. This allows learning better features, as well as a better correspondence search strategy that progressively exploits features from higher recall (robustness) to higher precision (spatial discrimination).
    }    
  \label{fig:teaser}
  \end{minipage}\hfill
\end{figure}

The current decade is witnessing as wide-ranging a revolution, brought about by the widespread use of deep neural networks. Yet there exist computer vision pipelines that, thanks to extensive engineering efforts, have proven impervious to end-to-end learned solutions. Despite some recent efforts~\cite{kendall2015iccv,vijayanarasimhan2017arxiv,brachmann2017cvpr}, deep learning solutions do not yet outperform or achieve similar generality as state-of-the-art methods on problems such as structure from motion (SfM)~\cite{wang2017icra} and object pose estimation~\cite{rad2017iccv}. Indeed, we see a consensus emerging that some of the systems employing interest point detectors and descriptors are here to stay, but it might instead be advantageous to leverage deep learning for their individual components.

Recently, a few convolutional neural network (CNN) architectures~\cite{yi2016eccv,choy2016nips,zeng2017cvpr,wang2017bmvc} have been proposed with the aim of learning strong geometric feature descriptors for matching images, and have yielded mixed results~\cite{schonberger2017cvpr,balntas2017cvpr}. We posit that the ability of CNNs to learn representation hierarchies, which has made them so valuable for many visual recognition tasks, becomes a hurdle when it comes to low-level geometric feature learning, unless specific design choices are made in training and inference to exploit that hierarchy. This paper presents such strategies for the problem of {\em dense geometric correspondence}.
 
Most recent works employ various metric learning losses and extract feature descriptors from the deepest layers~\cite{yi2016eccv,choy2016nips,zeng2017cvpr,wang2017bmvc}, with the expectation that the loss would yield good features right before the location of the loss layer. On the contrary, several studies~\cite{zeiler2014eccv,zhou2015iclr} suggest that deeper layers respond to high-level abstract concepts and are by design invariant to local transformations in the input image. However, shallower layers are found to be
more sensitive to local structure, which is not exploited by most deep-learning based approaches for geometric correspondence that use only deeper layers.
To address this, we propose a novel {\em hierarchical metric learning} approach that combines the best characteristics of various levels of feature hierarchies, to simultaneously achieve robustness and localization sensitivity.
Our framework is widely applicable, which we demonstrate through improved matching for interest points in both 2D and 3D data modalities, on KITTI Flow~\cite{menze2015cvpr} and 3DMatch~\cite{zeng2017cvpr} datasets, respectively.

Further, we leverage recent studies that highlight the importance of carefully marshaling the training process: (i) by deeply supervising~\cite{lee2015aistats,li2017cvpr} intermediate feature layers to learn task-relevant features, and (ii) on-the-fly hard negative mining~\cite{choy2016nips} that forces each iteration of training to achieve more. Finally, we exploit the intermediate activation maps generated within the CNN itself as a proxy for image pyramids traditionally used to enable coarse-to-fine matching ~\cite{czarnowski2017iccvw}. Thus, at test time, we employ a {\em hierarchical matching} framework, using deeper features to perform coarse matching that benefits from greater context and higher-level visual concepts, followed by a fine grained matching step that involves searching for shallower features. Figure \ref{fig:teaser} illustrates our proposed approach.


In summary, our contributions include:
\begin{tight_itemize}
\item We demonstrate that while in theory metric learning should produce good features irrespective of the layer the loss is applied to, in fact shallower
features are superior for high-precision geometric matching tasks, whereas deeper features help obtain greater recall.
\item We leverage deep supervision~\cite{lee2015aistats,li2017cvpr} for feature descriptor learning, while employing hard negative mining at multiple layers.
\item We propose a CNN-driven scheme for coarse-to-fine hierarchical matching, as an effective and principled replacement for conventional pyramid approaches.
\item We experimentally validate our ideas by comparing against state-of-the-art geometric matching approaches and feature fusion baselines,
 as well as perform an ablative analysis of our proposed solution. We evaluate for the tasks of 2D and 3D interest point matching and refinement, as well as optical flow, demonstrating state-of-the-art results and generalization ability.
\end{tight_itemize}

We review literature in Section~\ref{sec:related} and introduce our framework in Section~\ref{sec:method}.
We discuss experimental results in Section~\ref{sec:experiments}, concluding the paper in Section~\ref{sec:conclusion}.

%% file: related.tex
\section{Related Work}
\label{sec:related}

With the use of deep neural networks, many new ideas have emerged both pertaining to learned feature descriptors and directly learning networks for low-level vision tasks in an end-to-end fashion, which we review next.

{\bf Hand-Crafted Descriptors. } 
SIFT~\cite{lowe2004ijcv}, SURF~\cite{bay2006eccv}, BRISK~\cite{leutenegger2011brisk} were designed to complement high curvature point detectors, with \cite{lowe2004ijcv}
even proposing its own algorithm for such a detector. 
In fact, despite the interest in learned methods, they are still the state-of-the-art for precision~\cite{schonberger2017cvpr,balntas2017cvpr}, even if they are less effective in achieving high recall rates.


{\bf Learned Descriptors. }
While early work~\cite{weinzaepfel2013iccv,long2014nips,lin2016cvpr} leveraged intermediate activation maps of a CNN trained with an arbitrary loss
for keypoint matching, most recent methods rely on an explicit metric loss~\cite{zbontar2016jmlr,gadot2016cvpr,yi2016eccv,choy2016nips,zeng2017cvpr,yang2017iccv,zhang2017iccv}
to learn descriptors. The hidden assumption behind using contrastive or triplet loss at the final layer of a CNN is that this explicit loss will cause the relevant features
to emerge at the top of the feature hierarchy. But it has also been observed that early layers of the CNN are the ones that learn local geometric features \cite{zeiler2014eccv}. Thus, many of these works show superior performance to handcrafted descriptors on semantic matching tasks but often lag behind on geometric matching. 

{\bf Matching in 2D. }
LIFT~\cite{yi2016eccv} is a moderately deep architecture for end-to-end interest point detection and matching, which uses features at a single level of hierarchy and does not perform dense matching.
Universal Correspondence Network (UCN)~\cite{choy2016nips} combines a fully convolutional network in a Siamese setup, with a spatial transformer module~\cite{jaderberg2015nips} and 
contrastive loss~\cite{chopra2005cvpr} for dense correspondence, to achieve state-of-the-art on semantic matching tasks but not on geometric matching. Like them, we use GPU to speed up $k$-nearest neighbour for 
on-the-fly hard negative mining, albeit across multiple feature learning layers. Recently, AutoScaler~\cite{wang2017bmvc} explicitly applies a learned feature extractor on multiple scales of the input image. While this takes
 care of the issue that a deep layer may have an unnecessarily large receptive field when learning on the basis of contrastive loss, we argue that it is more elegant for the CNN 
 to ``look at the image'' at multiple scales, rather than separately process multiple scales.


{\bf Matching in 3D. }
Descriptors for matching in 3D voxel grid representations are learned by 3DMatch~\cite{zeng2017cvpr}, employing a Siamese 3D CNN setup on a 30x30x30 cm$^3$ voxel grid with a contrastive loss. It performs self-supervised learning by utilizing RGB-D scene reconstructions to obtain ground truth correspondence labels for training, outperforming a state-of-the-art hand-crafted descriptor~\cite{rusu2009icra}. 
Thus, 3DMatch provides an additional testbed to validate our ideas, where we report positive results from incorporating our hierarchical metric learning and matching into the approach.

{\bf Learned Optical Flow. }
Recent works achieve state-of-the-art results on optical flow by training CNNs in an end-to-end fashion~\cite{dosovitskiy2015iccv,ilg2017cvpr}, followed by Conditional Random Field (CRF) inference~\cite{revaud2015cvpr} to capture detailed boundaries.
We also demonstrate the efficacy of our matching on optical flow benchmarks. However, we do not use heavily engineered or end-to-end learning for minimizing flow metrics, rather we show that our matches along with an off-the-shelf interpolant~\cite{revaud2015cvpr} already yield strong results.

{\bf Deep Supervision. }
Recent works~\cite{lee2015aistats,li2017cvpr,li2018arxiv} suggest that providing explicit supervision to intermediate layers of a CNN can yield higher performance on unseen data, by regularizing the training process.
However, to the best of our knowledge, the idea has neither been tested on the task of keypoint matching nor had the learned intermediate features been evaluated. We do both in our work.

{\bf Image Pyramids and Hierarchical Fusion. }
Downsampling pyramids have been a steady fixture of computer vision for exploiting information across multiple scales~\cite{lucas1981ijcai}. Recently, many techniques have been developed for fusing features from different layers within a CNN and producing output at high resolution, \eg semantic segmentation~\cite{hariharan2015cvpr,ronneberger15miccai,pinheiro2016eccv,chen2017pami}, depth estimation~\cite{eigen14nips}, and optical flow~\cite{dosovitskiy2015iccv,ilg2017cvpr}. Inspired by~\cite{czarnowski2017iccvw} for image alignment, we argue that the growing receptive field in deep 
CNN layers~\cite{zeiler2014eccv} provides a natural way to parse an image at multiple scales. Thus, in our hierarchical
 matching scheme, we employ features extracted from a deeper layer with greater receptive field and higher-level 
 semantic notions~\cite{zhou2015iclr} for coarsely locating the corresponding point, followed by shallower features 
 for precise localization. We show gains in correspondence estimation by using our approach over prior feature fusion methods, \eg \cite{hariharan2015cvpr,pinheiro2016eccv}.

%% file: method.tex
\section{Method}
\label{sec:method}

\begin{figure}[t]
  \centering
    \includegraphics[width=1.0\linewidth]{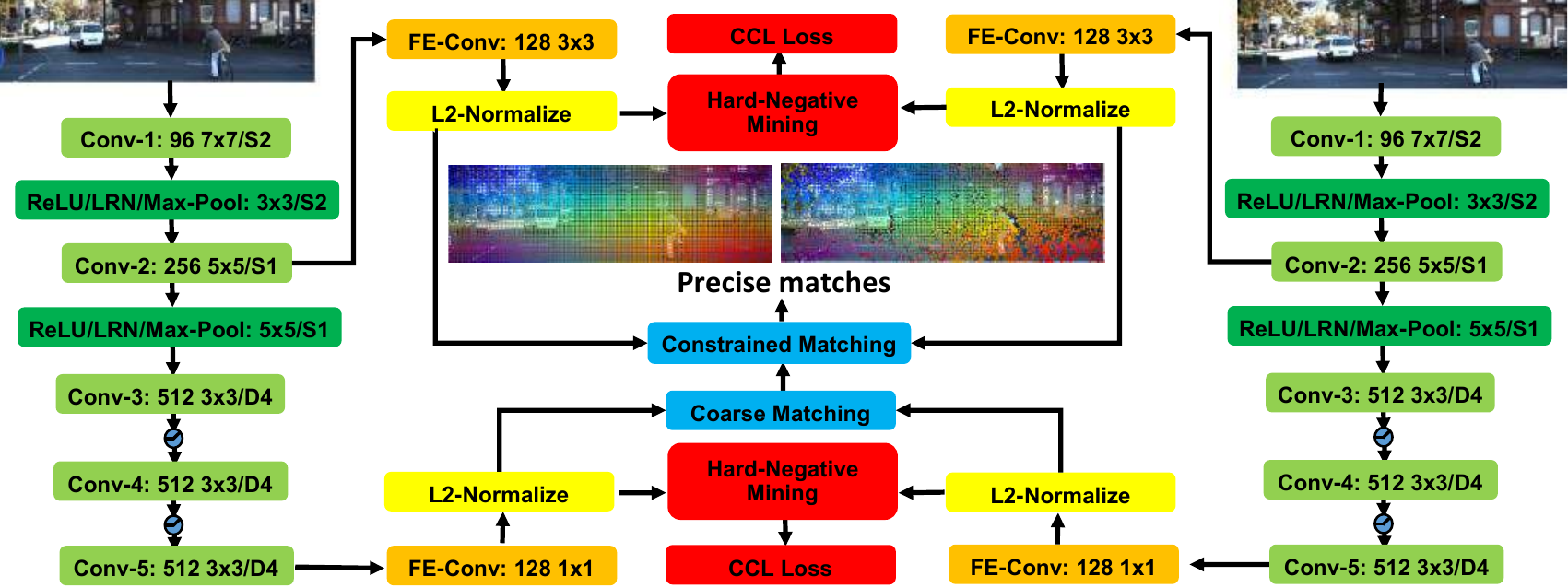}
  \caption{One instantiation of our proposed ideas. Note that the hard negative mining and CCL losses (red blocks) are relevant for training, and matching (blue blocks) for testing. Convolutional blocks (green) in the left and right Siamese branches share weights. \emph{`S'} and \emph{`D'} denote striding and dilation offsets.}
 \label{fig:arch}
\end{figure}

In the following, we first identify the general principles behind our framework, then propose concrete neural network architectures that realize them. In this section, we limit our discussion to models for 2D images. We detail and validate our ideas on the 3DMatch~\cite{zeng2017cvpr} architecture in Section~\ref{sec:3dmatchexp}.

\subsection{Hierarchical Metric Learning}
\label{sec:himl}
We follow the standard CNN-based metric learning setup proposed as the Siamese architecture~\cite{chopra2005cvpr}. This involves two Fully Convolutional Networks (FCN)~\cite{long2015cvpr} with tied weights, parsing two images of the same scene. We extract features out of the intermediate convolutional layer activation maps at the locations corresponding to the training points, and after normalization obtain their Euclidean distance. At training time, separate contrastive losses are applied to multiple levels in the feature hierarchy to encourage the network to learn embedding functions that minimizes the distance between the descriptors of matching points, while maximizing the distance between unmatched points.

{\bf Correspondence Contrastive Loss (CCL). }
 We borrow the
 correspondence contrastive loss formulation introduced
 in~\cite{choy2016nips}, and adapted from~\cite{chopra2005cvpr}. Here,
 $\phi_l^I(x)$ represents the feature extracted from the $l$-th feature
 level of the reference image $I$ at a pixel
 location $x$; similarly, $\phi_l^{I'}(x')$ represents the feature extracted from the $l$-th feature
 level of the target image $I'$ at a pixel
 location $x'$. Let $\mathcal D$ represent a dataset
 of triplets $(x,x',y)$, where $x$ is a location in the reference image $I$,
 $x'$ is a location in the target image $I'$, and $y \in \{0,1\}$ is $1$
 if and only if $(x,x')$ are a match.  Let $m$ be a margin parameter and $c$ be
 a window size. We define:
\begin{equation}
  \hat \phi_l^I(x) := \frac{\phi_l^I(x)}{\Vert \phi_l^I(x) \Vert_2}, \qquad
  d_l(x,x') := \Vert \hat \phi_l^I(x) - \hat \phi_l^{I'}(x') \Vert_2.
\end{equation}
Then, our training loss, $\mathcal L$, sums CCL losses over multiple levels $l$:
\begin{equation}
\mathcal L := \sum_{l=1}^L
  \sum_{(x,x',y)\in \mathcal D}
  y~.~d_l^2(x,x') + (1-y)~.~(\max(0, m - d_l(x,x')))^2.
  \label{eqn:ccl}
\end{equation}

{\bf Deep Supervision. }
Our rationale in applying CCL losses at multiple levels of the feature hierarchy is twofold. 
Recent studies~\cite{lee2015aistats,li2017cvpr} indicate that deep supervision contributes to improved regularization, by encouraging the network early on to learn task-relevant features. Secondly, both deep and shallow layers can be supervised for matching simultaneously within one network.

{\bf Hard Negative Mining. }
Since our training data includes only positive correspondences, we actively search for hard negative matches ``on-the-fly'' to speed up training and to leverage the latest instance of network weights. We adopt the approach of UCN~\cite{choy2016nips}, but in contrast to it, our hard negative mining happens independently for each of the feature levels being supervised.


{\bf Network Architectures. }
We visualize one specific instantiation of the above ideas in Figure~\ref{fig:arch}, adapting the VGG-M~\cite{chatfield2014bmvc} architecture for the task.
 We retain the first 5 convolutional layers, initializing them with weights pre-trained for ImageNet classification~\cite{russakovsky2015ijcv}. We use ideas from semantic segmentation literature~\cite{yu2016iclr,chen2017pami}
 to increase the resolution of the intermediate activation maps by (a) eliminating down-sampling in the second convolutional and pooling layers (setting their stride value to 1, down from 2) (b) increasing
 the pooling window size for the second layer from $3$x$3$ to $5$x$5$ and (c) dilating~\cite{yu2016iclr} the subsequent convolutional layers (\emph{conv3}, \emph{conv4} and \emph{conv5}) to retain their pretrained receptive fields.


At training, the network is provided with a pair of images and a set of point correspondences. The network is replicated in a Siamese scheme~\cite{chopra2005cvpr} during training
(with shared weights) where each sub-network processes one image from the pair; and thus after each feed-forward pass, we have 4 feature maps: 2 shallow ones and
2 deep ones, respectively from the second and fifth convolutional layers (\emph{conv2}, \emph{conv5}). We apply supervision after these same layers (\emph{conv2}, \emph{conv5}).

We also experiment with a GoogLeNet~\cite{szegedy2015cvpr} baseline as employed in UCN~\cite{choy2016nips}. Specifically, we augment the network with a $1$x$1$ convolutional layer and L2 normalization following the fourth convolutional block (\emph{inception}\textunderscore4a/output) for learning deep features, as in UCN. In addition, for learning shallow features, we augment the network with a $3$x$3$ convolutional layer right after the second convolutional layer (\emph{conv2}/$3$x$3$), followed by L2 normalization, but before the corresponding non-linear ReLU squashing function. We extract the shallow and deep feature maps based on the normalized outputs after the second convolutional layer \emph{conv2}/$3$x$3$ and the \emph{inception}\textunderscore4a/output layers respectively. We provide the detailed architecture of our GoogLeNet variant as supplementary material.


{\bf Network Training. }
We implement our system in Caffe~\cite{jia2014mm} and use ADAM~\cite{kingma2014iclr} to train our network for $50K$ iterations using a base learning rate of $10^{-3}$ on a P6000 GPU. Pre-trained layers are fine-tuned with a learning rate multiplier of 0.1 whereas the weights of the newly-added feature-extraction layers are randomly initialized using Xavier's method. We use a weight decay parameter of $10^{-4}$ and L2 weight regularization. During training, each batch consists of three randomly chosen image pairs and we randomly choose 1K positive correspondences from each pair. It takes the VGG-M variant of our system around 43 hours to train whereas it takes 30 hours to train our GoogLeNet-based variant.

\subsection{Hierarchical Matching}
We adapt and train our networks as described in the previous section, optimizing network weights for matching using features extracted from different layers. Yet, we find that features from different depths offer complementary capabilities as predicted by earlier works~\cite{zeiler2014eccv,zhou2015iclr} and confirmed by our empirical evaluation in Section~\ref{sec:experiments}. Specifically, features extracted from shallower layers obtain superior matching accuracies for smaller distance thresholds (precision), whereas those from deeper layers provide better accuracies for larger distance thresholds (recall). Such coarse-to-fine matching has been well-known in computer vision \cite{lucas1981ijcai}, however recent work highlights how employing CNN feature hierarchies for the task (at least in the context of image alignment~\cite{czarnowski2017iccvw}) is more robust.

To establish correspondences, we compare the deep and shallow features of the input images $I$ and $I'$ as follows. Assuming the shallow feature coordinates $p_s$ and the deep feature coordinates $p_d$ in the reference image $I$ are related by $p_d = p_s * 1/f$ with a scaling factor $f$, we first use the deep feature descriptor $\phi_{d}^{I}(p_d)$ in the reference image $I$ to find the point $p_d'$ in the target image $I'$ with $\phi_{d}^{I'}(p_d')$ closest to $\phi_{d}^{I}(p_d)$ with nearest neighbor search.\footnote{If $p_d$ is fractional, we use bilinear interpolation to compute $\phi_{d}^{I}(p_d)$.}  Next, we refine the location of $p_d'$ by searching within a circle of a radius of 32 pixels around $p_s' = p_d' * f$ (assuming input images have the same size, thus, $f' = f$) to find the point $\hat{p}_s'$ whose shallow feature descriptor $\phi_{s}^{I'}(\hat{p}_s')$ is closest to $\phi_{s}^{I}(p_s)$, forming a correspondence ($p_s$, $\hat{\p}_s'$).

Our proposed hierarchical matching is implemented on CUDA and run on a P6000 GPU, requiring an average of $8.41$ seconds to densely extract features and compute correspondences for a pair of input images of size $1242 \times 376$.



%% file: experiment.tex
\section{Experiments}
\label{sec:experiments}

In this section, we first benchmark our proposed method for 2D correspondence estimation against standard metric learning and matching approaches, 
feature fusion, as well as state-of-the-art learned and hand-crafted methods for extracting correspondences. Next, we show how our method 
for correspondence estimation can be applied for optical flow and compare it against recent optical flow methods. Finally, we 
incorporate our ideas in a state-of-the-art 3D fully convolutional network~\cite{zeng2017cvpr} and show improved performance.
In the following, we denote our 
method as \emph{HiLM}, which is short for \emph{Hi}erarchical metric \emph{L}earning and \emph{M}atching.

\input{validation}

\input{opticalflow}

\input{3dmatch}

%% file: validation.tex
\subsection{2D Correspondence Experiments}
\label{sec:validexp}

\begin{figure}[t]
  \centering
  \begin{minipage}[c]{0.60\textwidth}
    \includegraphics[width=0.98\linewidth]{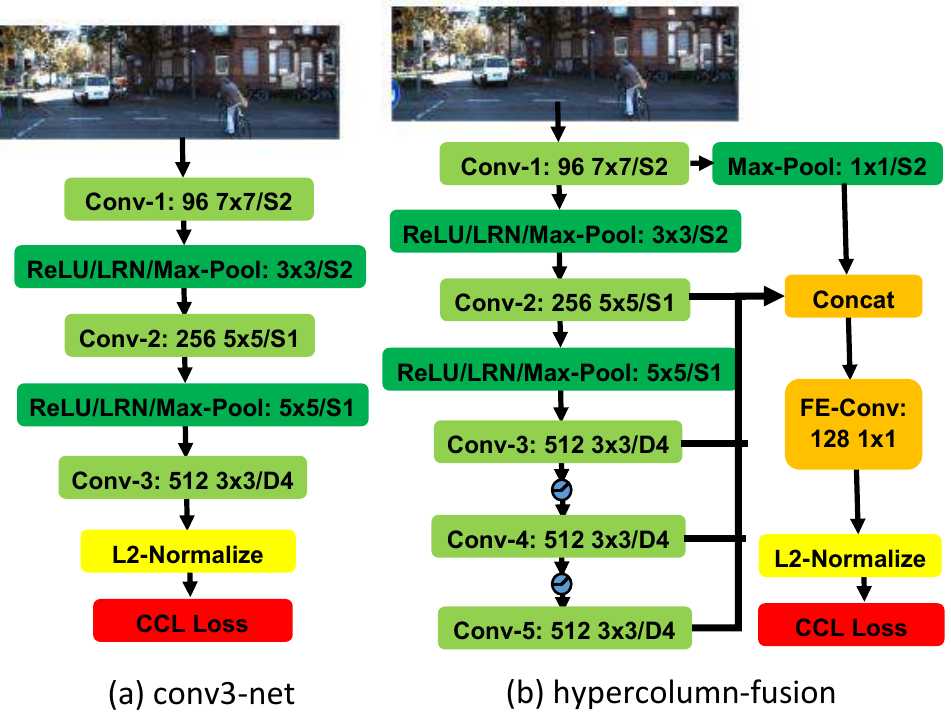}
  \end{minipage}\hfill
  \begin{minipage}[c]{0.40\textwidth}
    \caption{One Siamese branch of two for baseline architectures in our evaluation.
      The \emph{conv3-net} (a) is obtained by truncating all layers after VGG-M \emph{conv3} in Figure~\ref{fig:arch} and adding a convolutional layer, L2 normalization and CCL loss. Other \emph{convi-net} baselines are obtained similarly. The \axa{1} max pooling layer after \emph{conv1} in the \emph{hypercolumn-fusion} baseline (b) is added to down sample the \emph{conv1} feature map for valid concatenation with other feature maps. \emph{`S'} and \emph{`D'} denote striding and dilation offsets.}
    \label{fig:baselines}
  \end{minipage}\hfill
\end{figure}

\begin{figure}[t]
  \centering
  \begin{subfigure}{0.45\textwidth}
    \centering
    \includegraphics[width=1.0\linewidth]{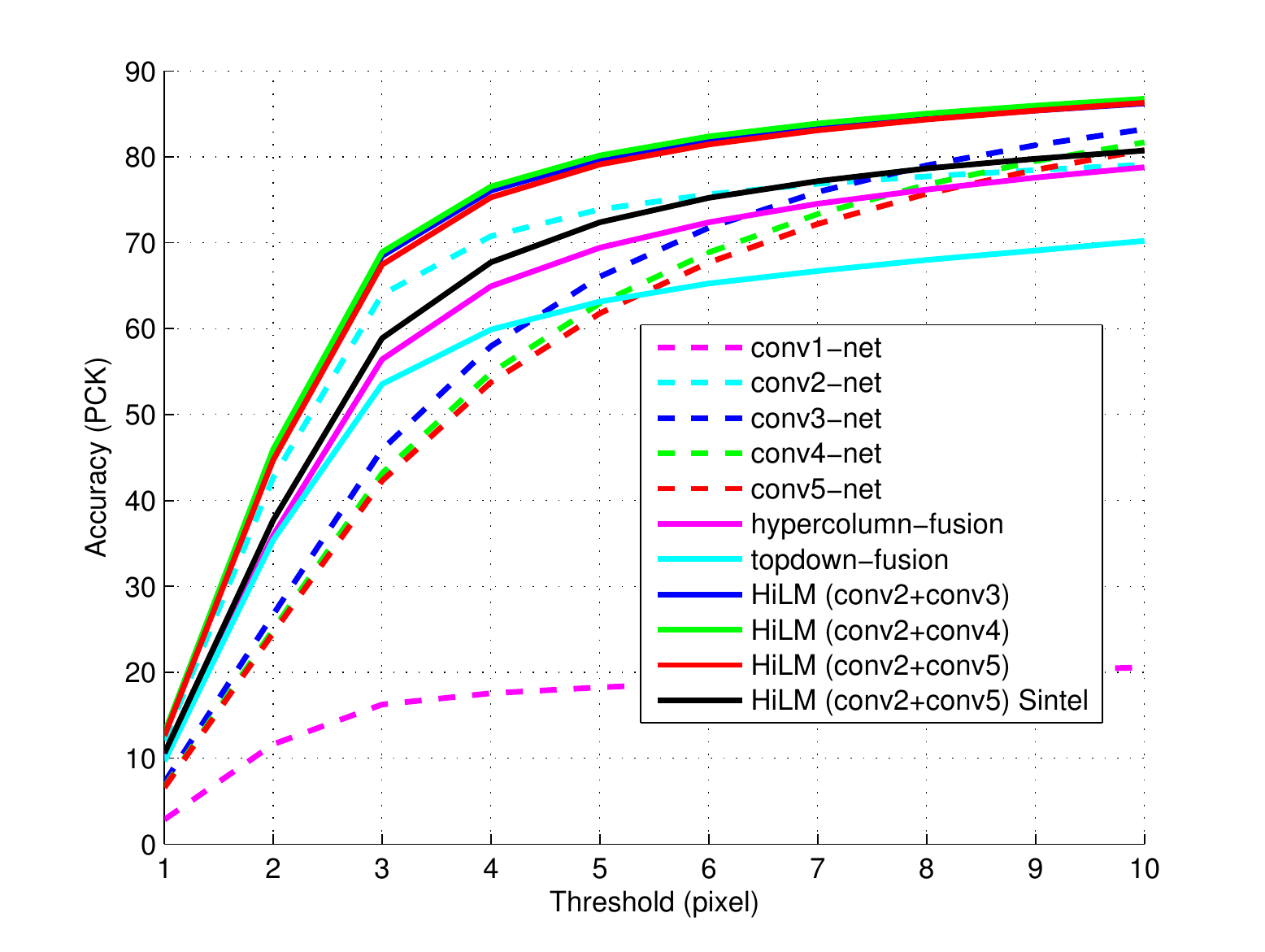}
    \caption{Accuracy over small thresholds}
    \label{fig:pck-small-kitti}
  \end{subfigure}
  \begin{subfigure}{0.45\textwidth}
    \centering
    \includegraphics[width=1.0\linewidth]{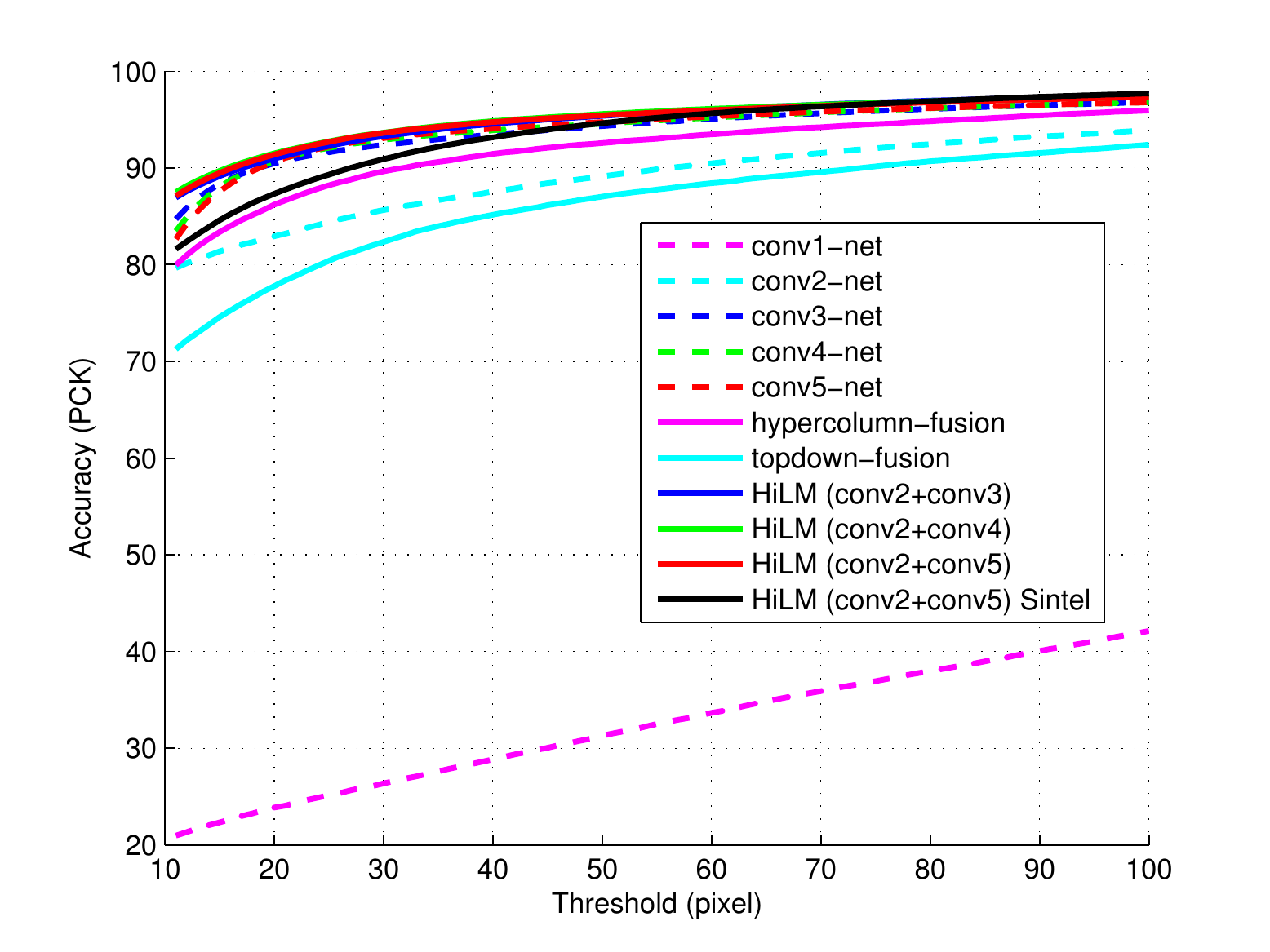}
    \caption{Accuracy over large thresholds}
    \label{fig:pck-large-kitti}
  \end{subfigure}
  \caption{Accuracy of different CNN-based methods for 2D correspondence estimation on KITTI Flow 2015.}
  \label{fig:pck-kitti}
\end{figure}

\begin{figure*}[t]
  \centering
  \begin{subfigure}{0.45\textwidth}
    \centering
    \includegraphics[width=1.0\linewidth]{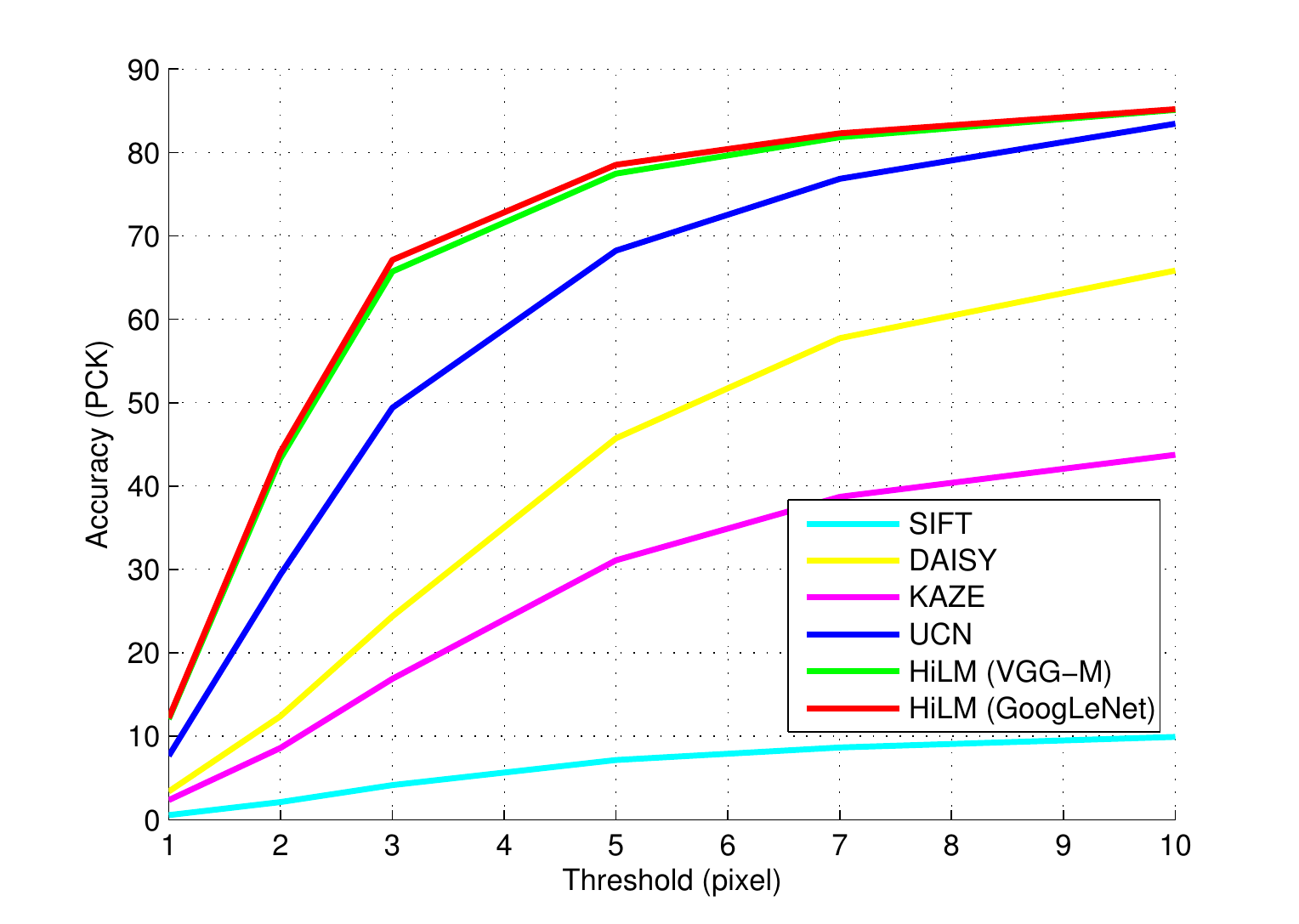}
    \caption{Accuracy over small thresholds}
    \label{fig:compare-small-kitti}
  \end{subfigure}
  \begin{subfigure}{0.45\textwidth}
    \centering
    \includegraphics[width=1.0\linewidth]{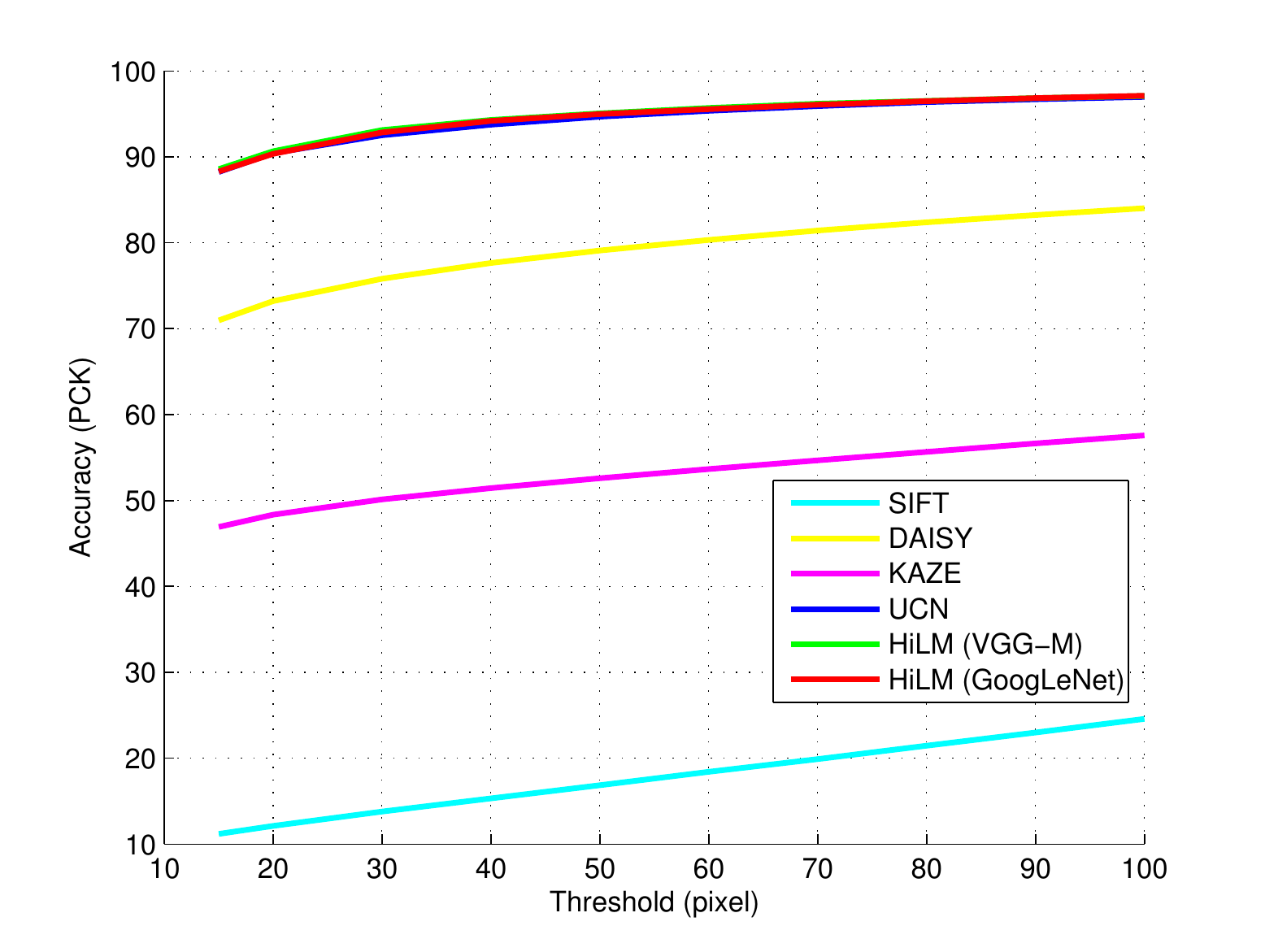}
    \caption{Accuracy over large thresholds}
    \label{fig:compare-large-kitti}
  \end{subfigure}
  \caption{Accuracy of CNN-based and hand-crafted methods for 2D correspondence estimation on KITTI Flow 2015.}
  \label{fig:compare-kitti}
\end{figure*}
We empirically evaluate our ideas against different approaches for dense correspondence estimation. We first consider
 metric learning and matching approaches based on feature sets extracted from a single convolutional layer~\footnote{LIFT~\cite{yi2016eccv} is not designed for dense matching and hence not included in our experiments. Note that LIFT also uses features from only a single convolutional layer.}, where we separately
 train five networks, based on the VGG-M baseline in Figure~\ref{fig:arch}. Each one of the five networks has a different depth and we refer to the $i$-th
 network by \emph{convi-net} to indicate that the network is truncated at the $i$-th convolutional layer (\emph{convi}), for $i \in {1, 2, ..., 5}$. We train a \emph{convi-net} network by adding a convolutional layer, L2 normalization, and CCL loss after the output of the last layer (\emph{convi}). Figure~\ref{fig:baselines} (a)
 shows one branch of the \emph{conv3-net} baseline as an example.

In addition, we also compare our method against two alternatives for fusing features from different layers inspired by ideas from semantic
segmentation~\cite{hariharan2015cvpr,pinheiro2016eccv}. One is \emph{hypercolumn-fusion} -- Figure~\ref{fig:baselines} (b), where feature sets from all layers
 (first through fifth) are concatenated for every interest point and  a set of \axa{1} convolution kernels are trained to fuse features
 before L2 normalization and CCL loss. Instead of upsampling deeper feature maps as in~\pcite{hariharan2015cvpr}, we extract deep features at higher resolution by setting the stride of multiple convolutional/pooling layers to $1$ while dilating the subsequent convolutions appropriately as
 shown in Figure~\ref{fig:baselines}. Another approach we consider is \emph{topdown-fusion}, where refinement modules similar to \cite{pinheiro2016eccv}
 are used to refine the top-level \emph{conv5} features gradually down the network by combining with lower-level features till \emph{conv2} (please see supplementary material for details).

We evaluate on KITTI Flow 2015~\cite{menze2015cvpr} where all networks are trained on $80\%$ of the image pairs and the remaining $20\%$ are used for evaluation.
For a fair comparison, we use the same train-test split for all methods and train each with 1K correspondences per image pair and for 50K iterations. During testing, we use the correspondences $\{(x_i, x'_i)\}$ in each image pair (obtained
using all non-occluded ground truth flows) for evaluation. Specifically, each method predicts a point $\hat{x}'_i$ in the target image that matches the point $x_i$ from the reference image $\forall i$.

{\bf Evaluation Metric. }
Following prior works~\cite{long2014nips,choy2016nips,wang2017bmvc}, we use Percentage of Correct Keypoints (PCK) as our evaluation metric. Given a pixel threshold $\theta$,
the PCK measures the percentage of predicted points $\hat{x}'_i$ that are within $\theta$ pixels from the ground truth corresponding point $x'_i$ (and so are
 considered as correct matches up to $\theta$ pixels).

{\bf Single-Layer and Feature Fusion Descriptors. }
 We plot PCK curves obtained for all methods under consideration in Figure~\ref{fig:pck-kitti} where we split the graph into sub-graphs based on the
 pixel threshold range. These plots reveal that, for smaller thresholds, shallower features (\eg \emph{conv2-net} with $73.89\%$ @ $5$ pixels) provide higher PCK than deeper ones (\eg \emph{conv5-net} with $61.78\%$ @ $5$ pixels),
 with the exception of \emph{conv1-net} which performs worst. Contrarily, deeper features have better performance for higher thresholds (\eg \emph{conv5-net} with $87.57\%$ versus \emph{conv2-net} with $81.36\%$ @ 15 pixels).\mycomment{ The graph
  suggests that deeper features are more suited for rough correspondence estimation whereas shallower features are more suited for obtaining more refined locations.}
  This suggests that, for best performance, one would need to utilize the shallower as well as deeper features produced by the network rather than just the
  output of the last layer.

The plot also indicates that while baseline approaches for fusing features improve the PCK for smaller
thresholds (\eg \emph{hypercolumn-fusion} with $69.41\%$ versus \emph{conv5-net} with $61.78\%$ @ $5$ pixels), they do not perform on par with the
simple \emph{conv2}-based features (\eg \emph{conv2-net} with $73.89\%$ @ $5$ pixels). 

Different variants of our full approach achieve the highest PCK for smaller thresholds (\eg HiLM (\emph{conv2}+\emph{conv4}) with $80.17\%$ @ 5 pixels), without losing accuracy for higher thresholds. \mycomment{By utilizing the power of the
shallow and deep features.} In fact, our method is able to outperform the \emph{conv2} features (\eg \emph{conv2-net} with $73.89\%$ @ $5$ pixels) although it uses them for refining the rough correspondences estimated
by the deeper layers. This is explained by the relative invariance of deeper features to local structure, which helps to avoid matching patches that have similar local
appearance but rather belong to different objects.

{\bf Generalization. }
We also perform experiments on cross-domain generalization ability. Specifically, we train HiLM (\emph{conv2}+\emph{conv5})
on MPI Sintel~\cite{sintel2012} and evaluate it on KITTI Flow 2015 as the previous experiment, plotting the result in Figure~\ref{fig:pck-kitti} (black curve).
As expected the Sintel model is subpar compared to the same model trained on KITTI ($72.37\%$ vs. $79.11\%$ @ $5$ pixels), however it outperforms
both \emph{hypercolumn-fusion} ($69.41\%$) and \emph{topdown-fusion} ($63.14\%$) trained on KITTI, across all PCK thresholds. Similar generalization results are obtained when cross-training with HPatches~\cite{balntas2017cvpr} (please see supplementary material for details).

{\bf Hand-Crafted Descriptors. }
We also compare the performance of (a) our HiLM (\emph{conv2}+\emph{conv5}, VGG-M), (b) a variant of our method based on
GoogLeNet/ UCN (described in Section~\ref{sec:method}), (c) the original UCN~\cite{choy2016nips}, and (d) the following hand-crafted descriptors:
SIFT~\cite{lowe2004ijcv}, KAZE~\cite{alcantarilla2012eccv}, DAISY~\cite{tola2010pami}. We use the same KITTI Flow 2015 evaluation set utilized in the previous
experiment. To evaluate hand-crafted approaches, we use them to compute the descriptors at test pixels in the reference image (for which ground truth correspondences are available) and match the resulting descriptors against the descriptors computed on the target image over a grid of 4 pixel spacing in both directions.

Figure~\ref{fig:compare-kitti} compares the resulting PCKs and shows that our HiLM (VGG-M) outperforms UCN~\cite{choy2016nips} for smaller thresholds (\eg HiLM (VGG-M) with $43.26\%$ versus UCN with $29.38\%$ @ 2 pixels). That difference in performance is not
the result of baseline shift since our GoogLeNet variant (same baseline network as UCN) has similar or slightly better performance
 compared to our VGG-M variant. The graph also indicates the relatively higher invariance of CNN-based descriptors to local structure that allows them to obtain a
 higher percentage of roughly-localized correspondences (\eg UCN with $83.42\%$, HiLM (VGG-M) with $85.08\%$, and HiLM (GoogLeNet) with $85.18\%$, all at 10 pixel threshold).

%% file: opticalflow.tex
\subsection{Optical Flow Experiments}
\label{sec:flowexp}

\begin{figure}[t]
  \centering
  \raisebox{-2.5cm}
  {
  \begin{minipage}[t]{0.63\textwidth}
    \includegraphics[width=0.98\linewidth]{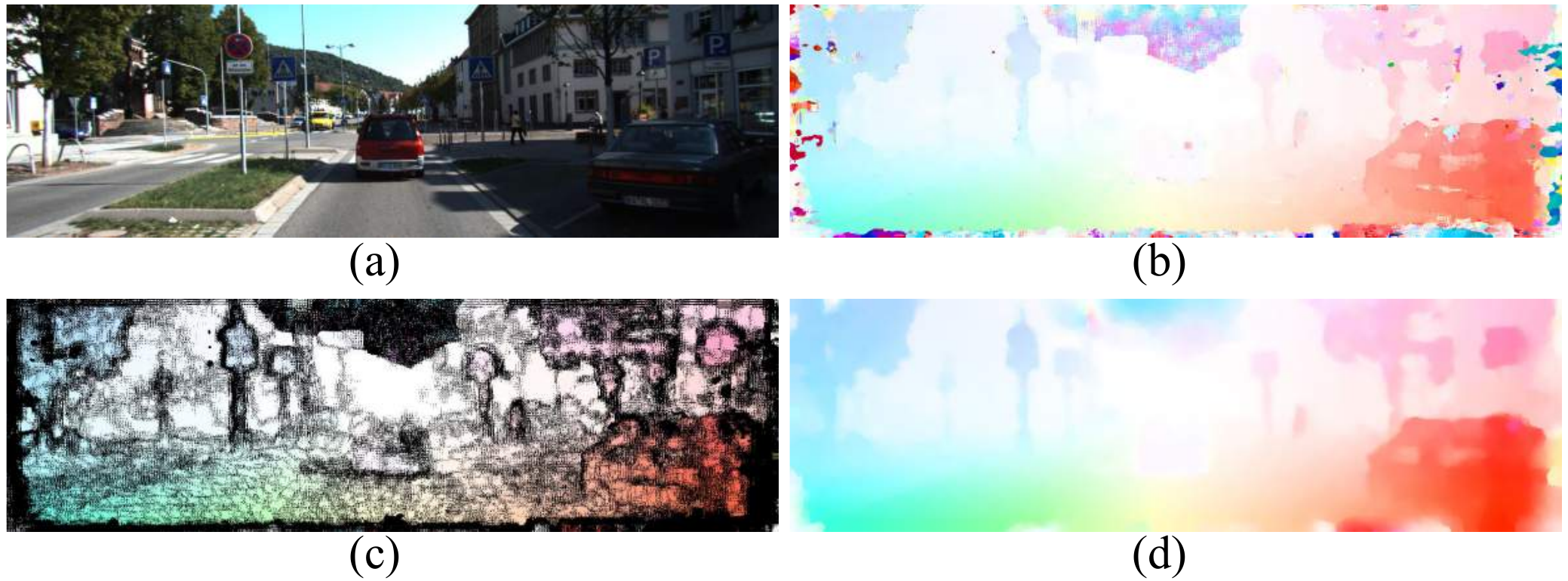}
  \end{minipage}\hfill
  }
  \begin{minipage}[t]{0.35\textwidth}
    \caption{Optical flow pipeline. (a) Input image. (b) Initial HiLM matches. (c) Filtered matches after consistency checks and motion constraints. (d) After interpolation using EpicFlow~\cite{revaud2015cvpr}.}
  \label{fig:flow_pipeline}
  \end{minipage}\hfill
\end{figure}

In this section, we demonstrate the application of our geometric correrspondences for obtaining optical flows. We
emphasize that the objective here is not to outperform methods that have been extensively engineered~\cite{bai2016eccv,sevillalara2016cvpr,ilg2017cvpr}
 for optical flows, including minimizing flow metric (end-point error) directly, \eg FlowNet2~\cite{ilg2017cvpr}.
 Yet, we consider it useful to garner insights from flow benchmarks since the tasks (\ie geometric correspondence and optical flow) are conceptually similar.

 \begin{table}[!!t]
   \centering
   \begin{minipage}[c]{0.50\textwidth}
     \small
     \raisebox{-2.8cm}{
     \begin{tabular}{| c | c | c | c |}
       \hline
       Method									& \emph{Fl-bg}		& \emph{Fl-fg}		& \emph{Fl-all} \\
       \hline
       FlowNet2~\cite{ilg2017cvpr}				& \textit{\underline{10.75}}\%	& \textbf{8.75}\%		& \textbf{10.41}\% \\
       SDF~\cite{bai2016eccv}					& \textbf{8.61}\%		& 26.69\%	& \textit{\underline{11.62}}\% \\
       SOF~\cite{sevillalara2016cvpr}			& 14.63\%	& 27.73\%	& 16.81\% \\
       CNN-HPM~\cite{bailer2017cvpr}			& 18.33\%	& 24.96\%	& 19.44\% \\
       \hdashline
       HiLM (Ours)									& 23.73\%	& \textit{\underline{21.79}}\%	& 23.41\% \\
       \hdashline
       SPM-BP~\cite{li2015iccv} 				& 24.06\%	& 24.97\%	& 24.21\% \\
       FullFlow~\cite{chen2016cvpr}				& 23.09\%	& 30.11\%	& 24.26\% \\
       AutoScaler~\cite{wang2017bmvc}			& 21.85\%	& 31.62\%	& 25.64\% \\
       EpicFlow~\cite{revaud2015cvpr}			& 25.81\%	& 33.56\%	& 27.10\% \\
       DeepFlow2~\cite{weinzaepfel2013iccv}		& 27.96\%	& 35.28\%	& 29.18\% \\
       PatchCollider~\cite{wang2016cvpr}		& 30.60\%	& 33.09\%	& 31.01\% \\
       \hline
     \end{tabular}
     }
   \end{minipage}\hfill
   \begin{minipage}[c]{0.45\textwidth}
     \caption{Quantitative results on KITTI Flow 2015. Following KITTI convention: \emph{`Fl-bl'}, \emph{`Fl-fg'}, and \emph{`Fl-all'} represent the outlier percentage on background pixels, foreground pixels and all pixels respectively. The methods are ranked by their \emph{`Fl-all'} errors. \textbf{Bold} numbers represent best results, while \textit{\underline{underlined}} numbers are second best ones. Note that FlowNet2~\cite{ilg2017cvpr} optimizes flow metric directly, while SDF~\cite{bai2016eccv} and SOF~\cite{sevillalara2016cvpr} require semantic knowledge.}
     \label{tab:quan_kitti}
   \end{minipage}\hfill
 \end{table}

\begin{figure*}[t]
  \centering
  \includegraphics[width=1.0\linewidth]{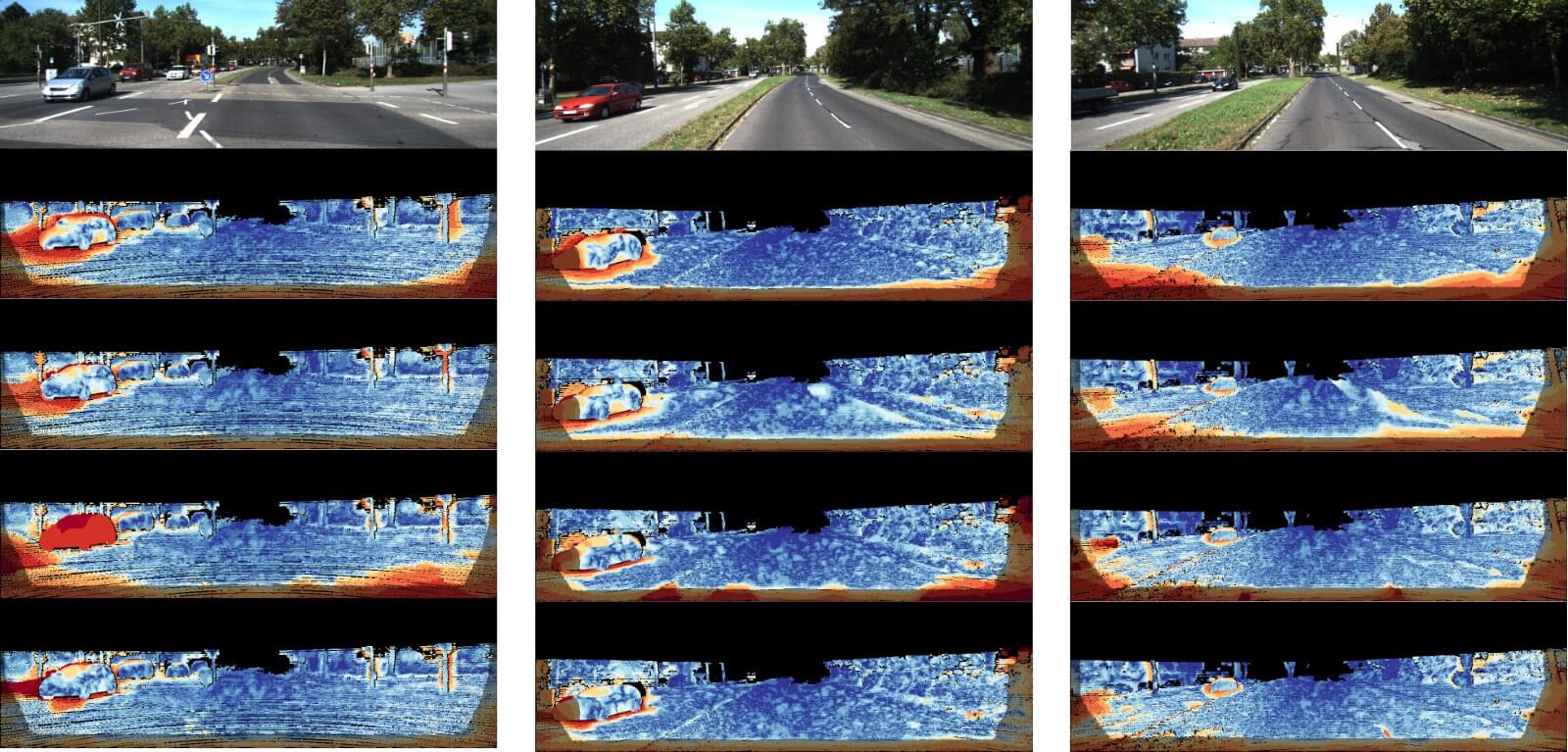}
  \caption{Qualitative results on KITTI Flow 2015. First row: input images. Second row: DeepFlow2~\cite{weinzaepfel2013iccv}. Third row: EpicFlow~\cite{revaud2015cvpr}. Forth row: SPM-BP~\cite{li2015iccv}. Fifth row: HiLM. Red colors mean high errors while blue colors mean low errors.}
  \label{fig:qual_kitti}
\end{figure*}

{\bf Network Architecture. }
For dense optical flow estimation, we leverage GoogLeNet \cite{szegedy2015cvpr} as our backbone architecture. However, at test time, we modify
the trained network to obtain dense per-pixel correspondences. To this end: (i) we set the stride to 1 in the first convolutional and pooling layers (\emph{conv1} and \emph{pool1}),
 (ii) we set the kernel size of the first pooling layer (\emph{pool1}) to 5 instead of 3, (iii) we set the dilation offset of the second convolutional layer (\emph{conv2})
 to 4, and (iv) we set the stride of the second pooling layer (\emph{pool2}) to 4. These changes allow us to obtain our shallow feature maps at the same resolution as the
 input images ($W$ x $H$) and the deep feature maps at $W/4$ x $H/4$, and to obtain dense per-pixel correspondences faster and with significantly fewer requirements on the GPU memory
as compared to an approach that would process the feature maps at full resolution through all layers of the network.

{\bf Procedure. }
We first extract and match feature descriptors for every pixel in the input images using our proposed method. These initial matches are usually contaminated by outliers or incorrect matches. Therefore, we follow the protocol of AutoScaler\cite{wang2017bmvc} for outlier removal. In particular, we enforce local motion
constraints using a window of $[-240,240] \textrm{x} [-240,240]$ and perform forward-backward consistency checks with a threshold of $0$ pixel. These filtered matches are then fed to EpicFlow~\cite{revaud2015cvpr} interpolation for producing the final optical flow output. Figure~\ref{fig:flow_pipeline} illustrates an example of this procedure.

{\bf Quantitative Evaluation. }
We tabulate our quantitative evaluation results on KITTI Flow 2015 in Table~\ref{tab:quan_kitti}. As mentioned earlier, our objective is not necessarily to obtain the best optical flow performance, rather we wish
to emphasize that we are able to provide high-quality interest point matches. In fact, many recent works~\cite{bai2016eccv,sevillalara2016cvpr} focus on embedding rich domain priors
at the level of explicit object classes into their models, which allows them to make good guesses when data is missing (\eg due to occlusions, truncations, homogenous surfaces). Yet, we are able to outperform several methods in our comparisons except~\cite{ilg2017cvpr} for foreground pixels (\ie by \emph{Fl-fg}, HiLM with $21.79\%$ versus other methods with $24.96$--$35.28\%$, excluding \cite{ilg2017cvpr} with $8.75\%$).
As expected, we do not get as good matches in regions of the image where relatively less structure is present (\eg background),
and for such regions methods~\cite{bai2016eccv,sevillalara2016cvpr} employing strong prior models have significant advantages. However, even on background regions, we are able to either beat or perform on par with most of our competitors (\ie by \emph{Fl-bg}, $23.73\%$ versus $18.33$--$30.60\%$), including machinery proposed for optical flows such as~\cite{weinzaepfel2013iccv,revaud2015cvpr,chen2016cvpr}.
Overall, we outperform 6 state-of-the-art methods evaluated in Table~\ref{tab:quan_kitti} (\ie by \emph{Fl-all}), including the multi-scale correspondence approach of~\cite{wang2017bmvc}.

{\bf Qualitative Evaluation. }
We plot some qualitative results in Figure \ref{fig:qual_kitti}, to contrast DeepFlow2~\cite{weinzaepfel2013iccv},
EpicFlow~\cite{revaud2015cvpr}, and SPM-BP~\cite{li2015iccv} against our method. As expected from the earlier discussion, we observe superior results for our method on the
image regions belonging to the vehicles, because of strong local structures, whereas for instance in first column (fourth row) SPM-BP~\cite{li2015iccv} entirely fails on the blue
car. We observe errors in the estimates of our method largely in regions which are occluded (surroundings of other cars) or truncated (lower portion of the images), where
the competing methods also have high errors.

%% file: 3dmatch.tex
\subsection{3D Correspondence Experiments}
\label{sec:3dmatchexp}

To demonstrate the generality of our contributions to different data modalities, we now consider an extension of our proposed method in Section~\ref{sec:method} to 3D correspondence estimation. In the following, we first present the details of our network architecture and then discuss the results of our quantitative evaluation.

{\bf Network Architecture. }
We use 3DMatch~\cite{zeng2017cvpr} as our baseline architecture. We insert two $3$x$3$x$3$ convolutional layers (stride of $2$ each)
 and one $5$x$5$x$5$ pooling layer (stride of $1$) after the second convolutional layer of 3DMatch to obtain a $512$-dimensional vector, which
 serves as the shallow feature descriptor. Our deep feature descriptor is computed after the eighth convolutional layer in the same
  manner as 3DMatch. Our hierarchical metric learning scheme again employs two CCL losses (Section ~\ref{sec:himl}) for learning shallow and deep feature descriptors
  simultaneously. We disable hard negative mining in this experiment to enable a fair comparison with 3DMatch.
   Our network is implemented in Marvin~\cite{xiao2015misc} and trained with stochastic gradient descent using a base learning rate
   of $10^{-3}$ for $137$K iterations on a TITAN XP GPU. We use pre-trained weights provided by 3DMatch to initialize the
   common layers in our network, which have a learning rate multiplier of $0.1$, whereas the weights of the newly added layers
   are initialized using Xavier's method and have a learning rate multiplier of $1.0$.
    We generate correspondence data for training using the same procedure as 3DMatch.

{\bf Protocol. }
    3DMatch evalutes classification accuracy of putative correspondences, using fixed
 keypoint locations and binary labels. Since our method enables refinement with shallow features and hence shifts
 hypothesized correspondence location in space, we define a protocol suitable to measure refinement performance. We employ PCK as our evaluation
  metric, similar to 2D experiments. We generate test data consisting of $10$K
  ground truth correspondences using the procedure of 3DMatch. We use a region of $30$x$30$x$30$ cm$^3$ centered on the reference
  keypoint (in the reference ``image'') following \cite{zeng2017cvpr} to compute the reference descriptor. This is matched against putative keypoints
  in a $60$x$60$x$60$ cm$^3$ region (in the target ``image''), to refine this coarse prior estimate\footnote{In fact, the ground truth keypoint correspondence lies at the center of this region, but this
  knowledge is not available to the method in any way.}. Specifically, we divide this region into
  subvolumes of $30$x$30$x$30$ cm$^3$ and employ our hierarchical matching approach to exhaustively search~\footnote{We use a sampling gap of 3 cm along all three dimensions in searching for subvolumes to reduce computational costs.} for the subvolume whose descriptor is most similar to the reference descriptor. In particular, once the coarse matching using deeper feature descriptors yields an approximate location in the $60$x$60$x$60$ cm$^3$ region, we constrain the refinement by shallow feature descriptors to a search radius of $15$ cm around the approximate location returned from the coarse matching.

{\bf Quantitative Evaluation. }
We compare our complete framework, namely, HiLM (\emph{conv2}+\emph{conv8}) against variants which are trained with hierarchical metric loss but rely either on deep or shallow features for matching (HiL (\emph{conv8}) and HiL (\emph{conv2}), respectively), and 3DMatch which use only deep features.
 Figure~\ref{fig:pck-3dmatch} shows the PCK curves of all competing methods computed over 10K test correspondences generated by the procedure
 of 3DMatch. From the results, our shallow features trained with hierarchical metric learning are able to outperform their deep counterparts for
  most PCK thresholds (\eg HiL (\emph{conv2}) with $21.50\%$ versus HiL (\emph{conv8}) with $20.78\%$ @ 9 cm). By utilizing both deep and shallow features, our complete framework achieves higher PCK numbers than its variants and outperforms 3DMatch across all PCK thresholds (\eg HiLM (\emph{conv2}+\emph{conv8}) with $24.36\%$ versus 3DMatch with $22.04\%$ @ 9 cm).

\begin{figure*}[t]
  \centering
  \begin{subfigure}{0.45\textwidth}
    \centering
    \includegraphics[width=1.0\linewidth]{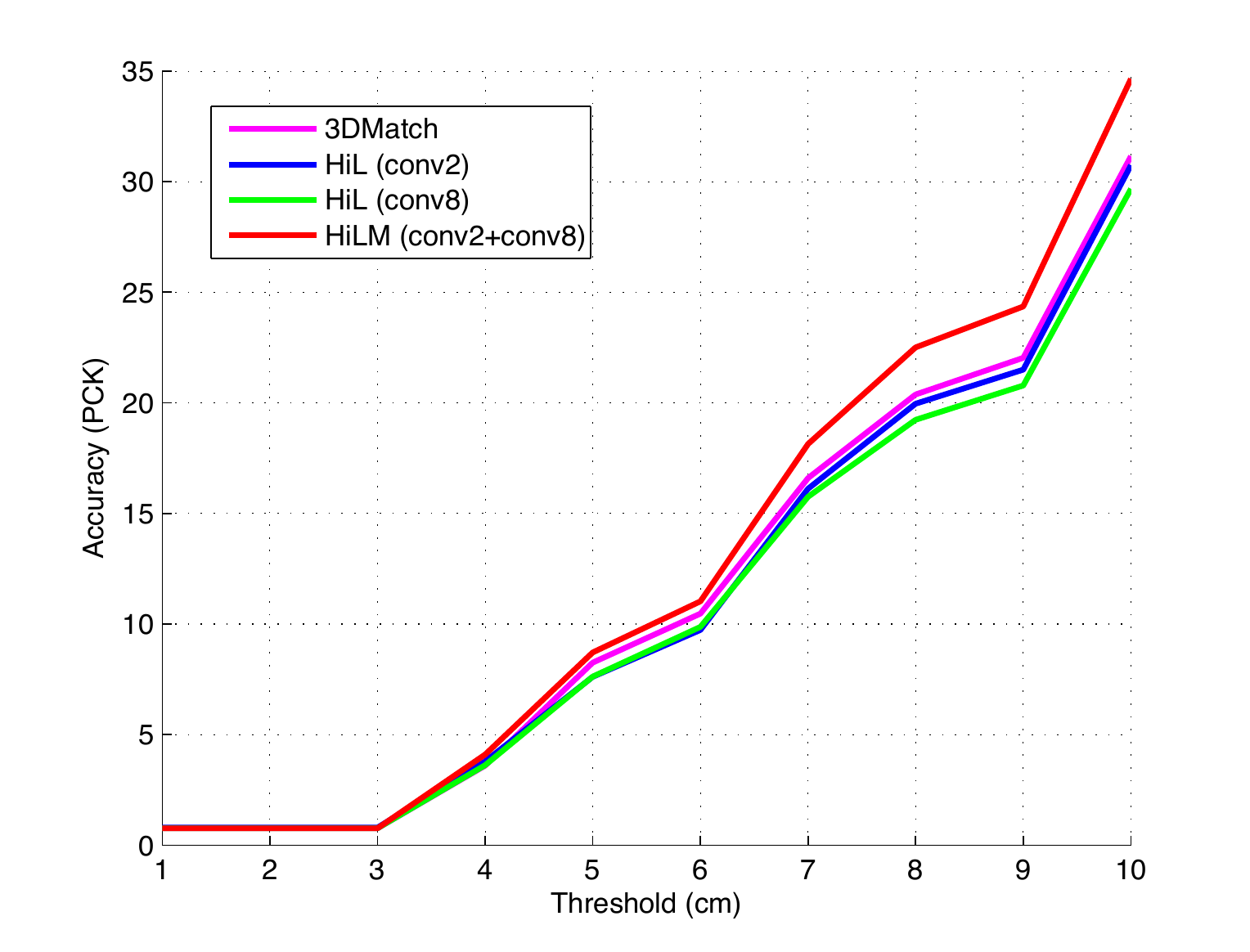}
    \caption{Accuracy over small thresholds}
    \label{fig:pck-small-3dmatch}
  \end{subfigure}
  \begin{subfigure}{0.45\textwidth}
    \centering
    \includegraphics[width=1.0\linewidth]{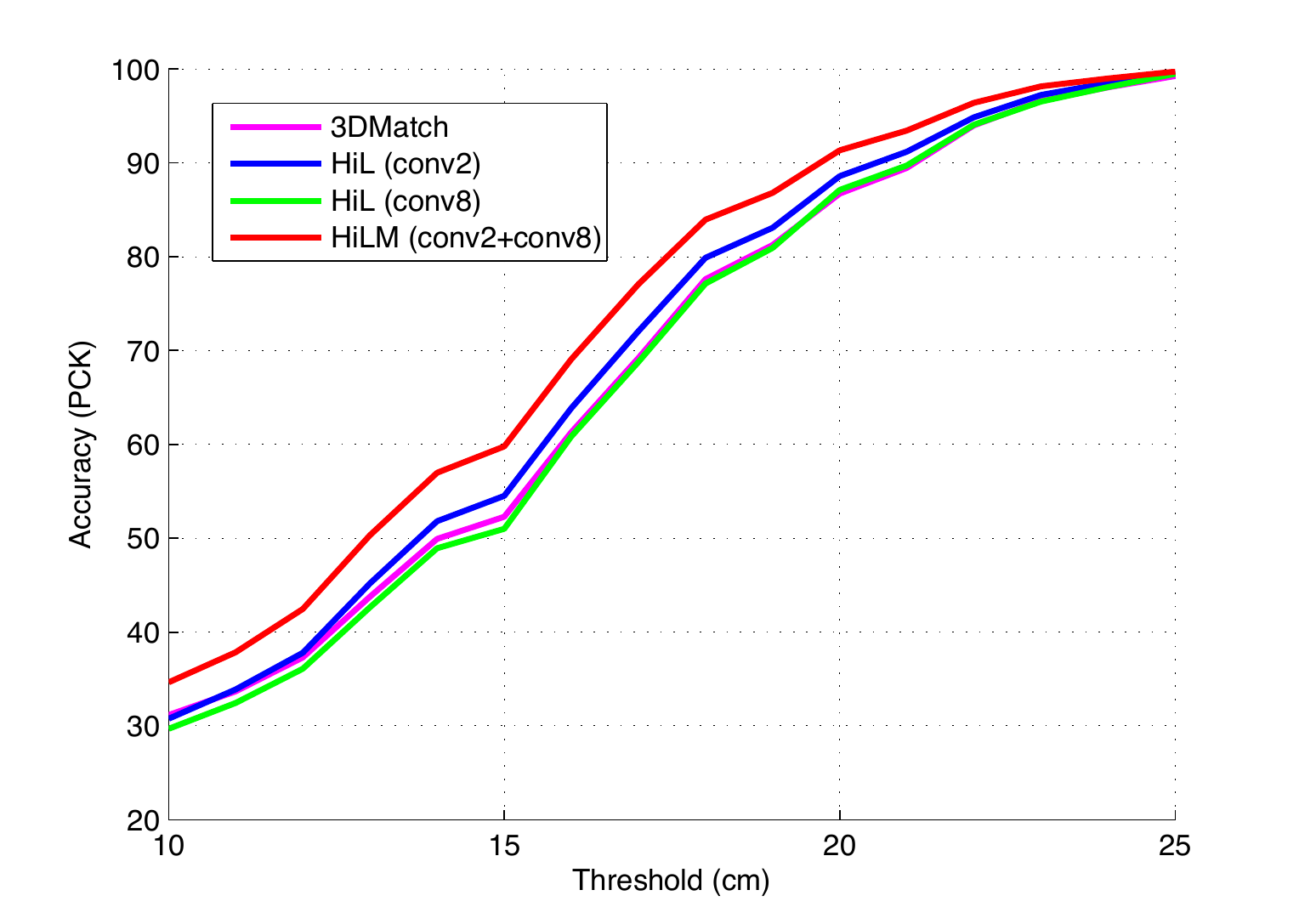}
    \caption{Accuracy over large thresholds}
    \label{fig:pck-large-3dmatch}
  \end{subfigure}
  \caption{Accuracy of different CNN-based methods for 3D correspondence estimation.}
  \label{fig:pck-3dmatch}
\end{figure*}

\mycomment{
\begin{figure}[t]
  \centering
  \begin{minipage}[c]{0.8\textwidth}
    \includegraphics[width=.495\linewidth]{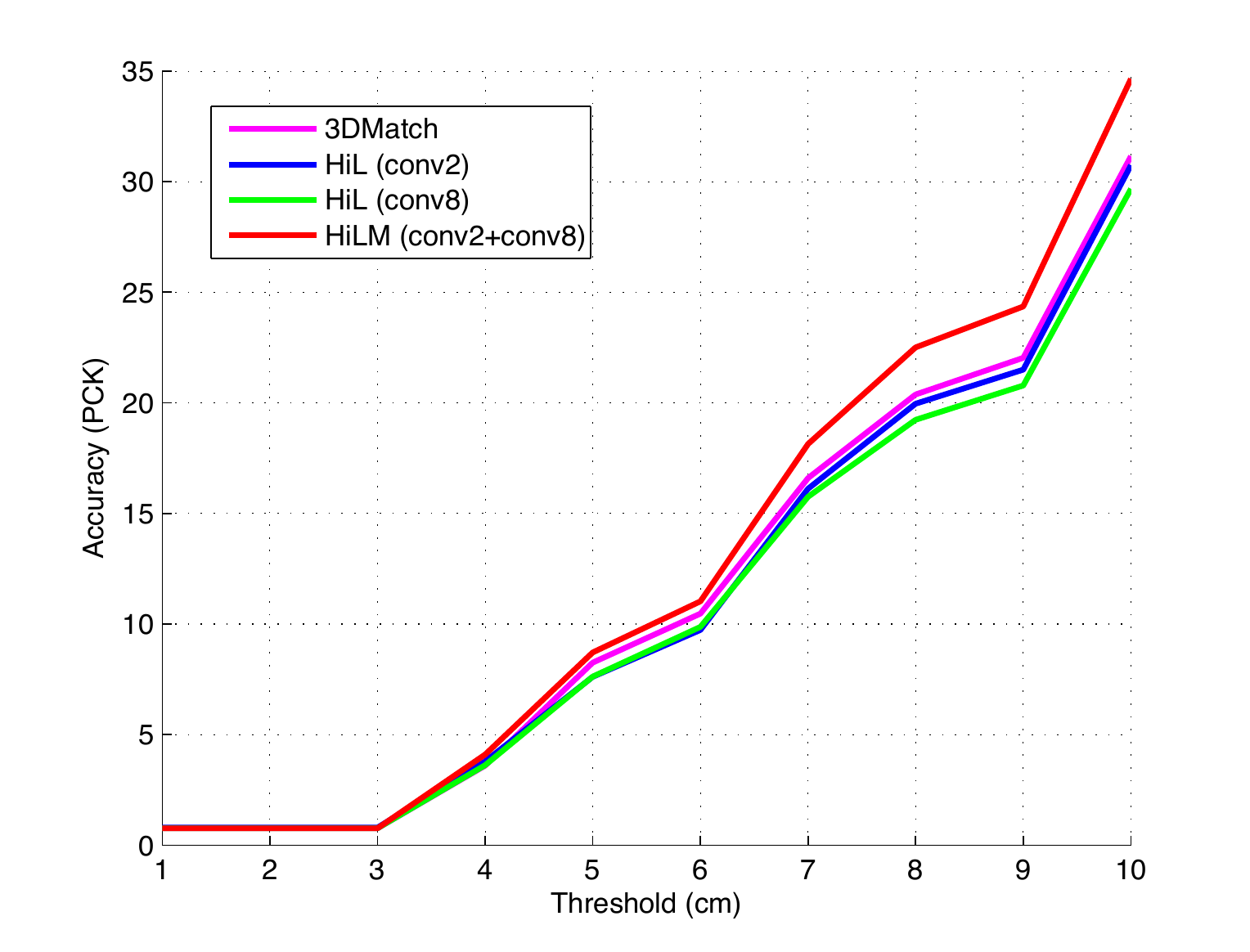}
    \includegraphics[width=.495\linewidth]{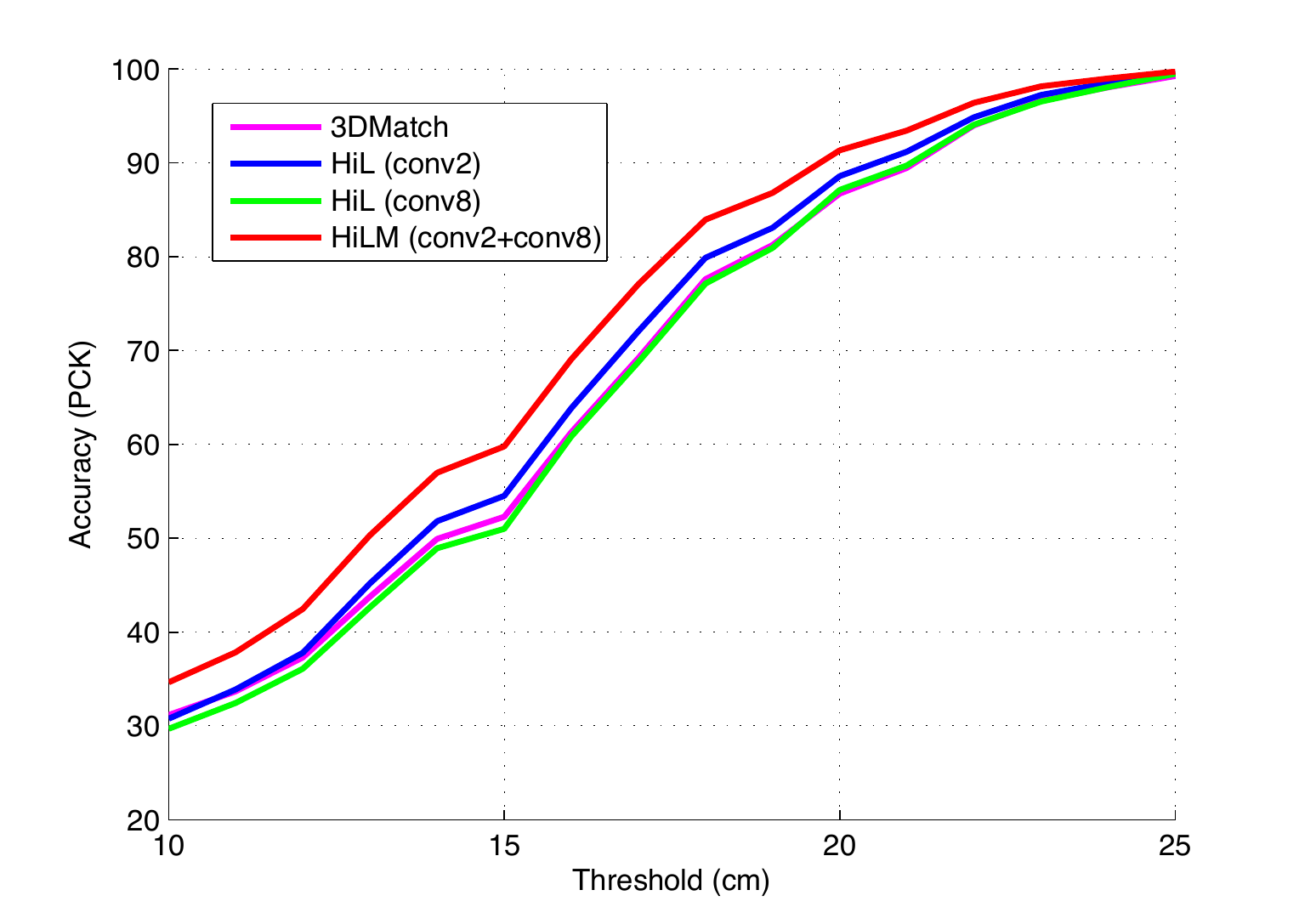}
  \end{minipage}
  \begin{minipage}{0.19\textwidth}
    \caption{Accuracy of different CNN-based methods for 3D correspondence estimation over small thresholds (left) and large ones (right).}
    \label{fig:pck-3dmatch}
  \end{minipage}
\end{figure}
}

%% file: conclusion.tex
\section{Conclusion and Future Work}
\label{sec:conclusion}
We draw inspiration from recent studies~\cite{zeiler2014eccv,zhou2015iclr} as well as conventional intuitions about CNN architectures to enhance learned representations for dense 2D and 3D geometric matching. Convolutional network architectures naturally learn hierarchies of features, thus, a contrastive loss applied at a deep layer will return features that are less sensitive to local image structure. We propose to remedy this by employing features at multiple levels of the feature hierarchy for interest point description. Further, we leverage recent ideas in deep supervision to explicitly obtain task-relevant features at intermediate layers. Finally, we exploit the receptive field growth for increasing layer depths as a proxy to replace conventional coarse-to-fine image pyramid approaches for matching. We thoroughly evaluate these ideas realized as concrete network architectures, on challenging benchmark datasets. Our evaluation on the task of explicit keypoint matching outperforms hand-crafted descriptors, a state-of-the-art descriptor learning approach~\cite{choy2016nips}, as well as various ablative baselines including hypercolumn-fusion and topdown-fusion. Further, an evaluation for optical flow computation outperforms several competing methods even without extensive engineering or leveraging higher-level semantic scene understanding. Finally, augmenting a recent 3D descriptor learning framework~\cite{zeng2017cvpr} with our ideas yields performance improvements, hinting at wider applicability. Our future work will explore applications of our correspondences, such as flexible ground modeling~\cite{lee2015icra,dhiman2016cvpr,ansari2018arxiv} and geometric registration~\cite{choi2015cvpr,zeng2017cvpr}.